\documentclass[letterpaper, 10 pt, conference]{ieeeconf}  

\IEEEoverridecommandlockouts                              

\usepackage{graphicx}
\usepackage{multirow}
\usepackage{marvosym} 
\usepackage{threeparttable}
\usepackage{arydshln}
\usepackage{amsmath}
\usepackage{amssymb}

\usepackage[colorlinks,linkcolor=blue,
anchorcolor=blue,
citecolor=green
]{hyperref} 

\title{\LARGE \bf
PKU-GoodsAD: A Supermarket Goods Dataset for Unsupervised Anomaly Detection and Segmentation
}

\author{Jian Zhang, Runwei Ding$^{*}$, Miaoju Ban, Ge Yang  
\thanks{This work is supported by National key R\&D program of China (2018YFB1308602, 2018YFB1308600), and Science and Technology Plan of Shenzhen (JCYJ20200109140410340).}
\thanks{$^{*}$Corresponding author. The authors are with the Key Laboratory of Machine Perception, Shenzhen Graduate School, Peking University, Beijing 100871, China. Email: {\{zhangjian, miaoju.ban\}@stu.pku.edu.cn, dingrunwei@pku.edu.cn, yangge@pkusz.edu.cn}}}

\begin{document}

\maketitle
\thispagestyle{empty}
\pagestyle{empty}

\begin{abstract}
Visual anomaly detection is essential and commonly used for many tasks in the field of computer vision. Recent anomaly detection datasets mainly focus on industrial automated inspection, medical image analysis and video surveillance. In order to broaden the application and research of anomaly detection in unmanned supermarkets and smart manufacturing, we introduce the supermarket goods anomaly detection (GoodsAD) dataset. It contains 6124 high-resolution images of 484 different appearance goods divided into 6 categories. Each category contains several common different types of anomalies such as deformation, surface damage and opened. Anomalies contain both texture changes and structural changes. It follows the unsupervised setting and only normal (defect-free) images are used for training. Pixel-precise ground truth regions are provided for all anomalies. Moreover, we also conduct a thorough evaluation of current state-of-the-art unsupervised anomaly detection methods. This initial benchmark indicates that some methods which perform well on the industrial anomaly detection dataset (e.g., MVTec AD), show poor performance on our dataset. This is a comprehensive, multi-object dataset for supermarket goods anomaly detection that focuses on real-world applications.\\



\end{abstract}


\section{INTRODUCTION}
Anomaly areas are regions that differ from normal areas. While humans can easily identify anomaly areas on the surface of objects based on their learned knowledge, it is challenging for machines to do the same.

Visual Anomaly detection (VAD) is one of the essential applications in the field of computer vision, which aims to classify and locate anomaly regions. Currently, anomaly detection algorithms are widely used in various fields such as industrial quality inspection, medical diagnosis, and intelligent surveillance. Specifically, in the field of industrial quality inspection, anomaly detection can be used to detect defects on the surface of industrial products. In the field of medical diagnosis, it can be used to detect lesions on the surface of organs. In the field of intelligent surveillance, it can be used to detect the occurrence of anomalous events. Therefore, it has broad application prospects and research significance. Due to the scarcity of anomalous data, unsupervised anomaly detection algorithms have drawn much attention in research. Their goal is to train models only using a large amount of easily obtainable normal samples, enabling the models to differentiate anomalous samples. At present, unsupervised anomaly detection algorithms can be divided into three categories: those based on pre-trained models, those based on pseudo anomaly generation, and those based on generative models. The first category uses a model that has learned features of normal samples from the ImageNet dataset to distinguish anomalous samples. The second category generates pseudo anomalies that resemble real anomalies during training, thereby transforming the unsupervised paradigm into a supervised one. The third category trains the model to fit the distribution of normal samples and distinguishes anomalies by calculating the distance between the distributions of anomalous and normal samples during training. Due to their ability to distinguish anomalies without using anomalous samples, these unsupervised anomaly detection methods have gained increasing attention and achieved remarkable results in various academic conferences.
\begin{figure}[t]
	\centering
	\includegraphics[width=1.0\columnwidth]{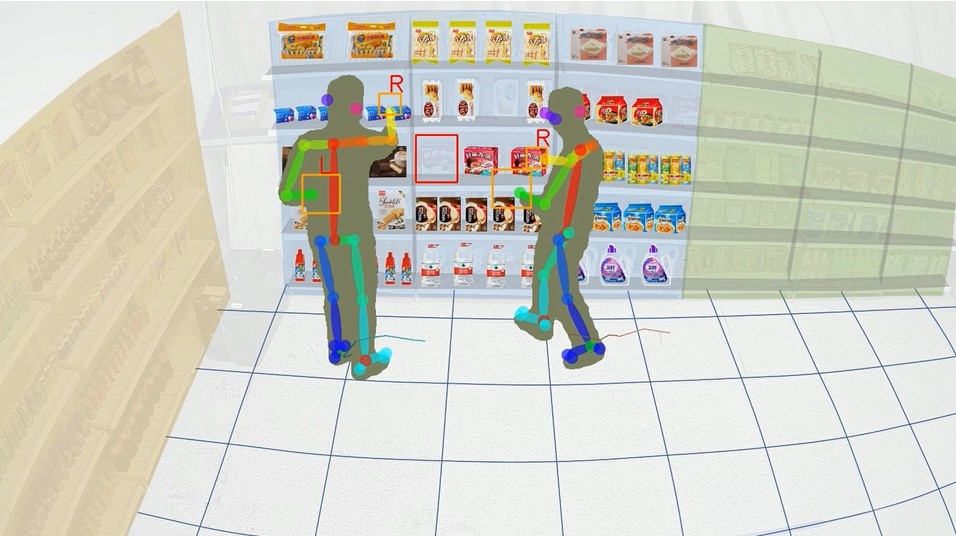}
	\caption{Virtual scene of unmanned supermarket shopping based on computer vision technology.}
	\label{fig:mkt}
\end{figure}
\par
However, most existing anomaly detection datasets are limited and mainly concentrated in industrial quality inspection, medical diagnosis, and intelligent monitoring fields. The diversity of datasets in these fields is also limited. Currently, widely used unsupervised anomaly detection datasets include MVTec AD\cite{bergmann2019mvtec} in industrial quality inspection, Chest X-ray\cite{kermany2018identifying} in medical diagnosis, and ShanghaiTech\cite{luo2017revisit} in intelligent monitoring. Due to the high-speed development of this field and the high cost of dataset construction, the performance of existing datasets has approached saturation, limiting the development of anomaly detection. With the continuous development of intelligence, unmanned supermarkets (Fig. \ref{fig:mkt}) have entered people's lives. Although the shopping process does not require human intervention, detecting and replacing damaged goods in unmanned supermarkets often requires a large amount of manpower. The demand for anomaly goods detection in supermarkets is increasing day by day, but there is currently a lack of large-scale anomaly goods datasets. Therefore, establishing an unsupervised anomaly goods dataset has significant research value and application prospects.

Based on this, we collected a large number of normal and anomaly goods sample images in a real unmanned supermarket application scenario and performed pixel-level anomaly annotation, creatively establishing the first goods anomaly detection (GoodsAD) dataset\footnote{\tt https://github.com/jianzhang96/GoodsAD} in the field of artificial intelligence. The dataset contains a total of six goods categories, including boxed cigarettes, bottled drinks, canned drinks, bottled foods, boxed foods, and packaged foods. Each goods has multiple types of anomalies, totalling 8 different types. The dataset includes 6,124 images, with 4,464 images of normal goods and 1,660 images of anomaly goods. The resolution of the images is 3000 × 3000. In the experiment, we selected 3,136 normal images as the training set and used the remaining 2,988 normal and anomaly images as the test set. In addition, we also tested the goods dataset on current state-of-the-art (SOTA) unsupervised anomaly detection methods and compared the performance of various methods. Our contribution is twofold:
\begin{itemize}
    \item We creatively established the first unsupervised anomaly detection goods dataset in the field of artificial intelligence, which is used to classify and locate anomaly areas on the surface of goods, increase the diversity of data in the anomaly detection field, and promote the development of unmanned supermarkets.
    \item Extensive experiments are conducted on the established goods dataset using current unsupervised anomaly detection methods, laying the foundation for subsequent anomaly detection work and promoting the performance improvement of related algorithms.
    
\end{itemize}

\section{RELATED WORK}
Some previous anomaly detection methods made experiments on image classification datasets such as MNIST and CIFAR10. They assume that a certain category of the dataset is normal and the rest is anomalous. For the application of visual anomaly detection, industrial vision \cite{ban2022pdd}, medical image analysis and video anomaly detection \cite{bozcan2021gridnet} are fields of great concern. Table \ref{tab:dataset} shows commonly used datasets for visual anomaly detection. In the field of medical images, there are datasets for anomaly detection such as Chest X-ray \cite{kermany2018identifying} and CheXpert \cite{irvin2019chexpert}. ShanghaiTech \cite{luo2017revisit} and Avenue \cite{lu2013abnormal} are two commonly used datasets for video anomaly detection. Some anomaly detection datasets \cite{xie2023iad} in the industry field have been proposed in recent years. These datasets all provide pixel-level annotations. DAGM \cite{wieler2007weakly} and NEU-SDD \cite{song2013noise} are early datasets. DAGM contains 10 types of texture images with artificial defects. NEU-SDD contains 6 kinds of typical surface defects of the hot-rolled steel strip. MTD \cite{huang2020surface} includes 6 types of defects on the surface of magnetic tiles. This dataset is somewhat difficult because the contrast of some defects and background is low. MSD \cite{zhang2022fdsnet} dataset contains three types of defects in mobile phone screens. These four datasets all follow the supervised learning setting.\par
\begin{table}[t]
\setlength\tabcolsep{1pt}
\centering
\caption{Datasets for Visual anomaly detection}
\begin{tabular}{c|c|cc|cc|cc|c}
\hline
\multirow{2}{*}{Field} & \multirow{2}{*}{Dataset} & \multicolumn{2}{c|}{Samples} & \multicolumn{2}{c|}{Classes} & \multicolumn{2}{c|}{Resolution} & \multirow{2}{*}{Year} \\ 
             & & \scriptsize{Normal}       & \scriptsize{Anomaly}      & \scriptsize{Object}    & \scriptsize{Anomalies}    & \scriptsize{Min}           & \scriptsize{Max}            \\ \hline

\multirow{9}{*}{\rotatebox{90}{Industry}} & VisA \cite{zou2022spot}        &  10621            &     1200         &     12      &   78              &       960        &      1562        & 2022  \\
&MSD \cite{zhang2022fdsnet} & 0 & 1200 & 1 & 3 & 1080 & 1920 & 2022 \\
&MPDD \cite{jezek2021deep}        &     1064         &     282         &      6     &  12               &    1024           &    1024      & 2021      \\
&BTAD \cite{mishra2021vt}        &    2250          &       580       &     3      &  3               &      600         &   1600    & 2021         \\
&KSDD2 \cite{tabernik2020segmentation} &       2979       &       356       &     1      &        5         &  230             &     630     & 2021      \\
&MVTec AD \cite{bergmann2019mvtec}     & 4096         & 1258         & 15        & 73              & 700           & 1024      & 2019     \\
&MTD \cite{huang2020surface}         &    952          &      392        &      1     &    5             &     105          & 516   & 2018 \\  
&NEU-SDD \cite{song2013noise}      &    0          &    1800          &      1     &       6          &        300       &        300     & 2013   \\
&DAGM \cite{wieler2007weakly}         &     15000         &     2100         &      10     &         10        &   512            &        512     & 2007   \\   \hline
\multirow{4}{*}{\rotatebox{90}{Medicine}}
& VinDr-CXR \cite{nguyen2022vindr} & 12658 & 5342 &1 & 28 & 2394 & 2788 & 2022\\
& CheXpert \cite{irvin2019chexpert} &158976 &65240 &1 &14 &320 &320 &2019 \\
& Chest X-ray \cite{kermany2018identifying} &81315 &30805 &1 &14 &1024 &1024 &2017 \\
&  MURA \cite{rajpurkar2017mura} & 9045 & 5818 & 7& 4 & 1500 & 2000 & 2017\\ \hline
\multirow{3}{*}{\rotatebox{90}{Video}}
& Avenue \cite{lu2013abnormal} &10 &6 &1 &5 &360 &640 &2018 \\
& UCF-Crime \cite{sultani2018real} & 950 & 950 & 1 & 13& 64& 64& 2018\\
& ShanghaiTech \cite{luo2017revisit} &661 &537 &13 &11 &768 &2448 &2016 \\
\hline
\end{tabular}
\label{tab:dataset}
\end{table}

\begin{figure*}[t]
	\centering
	\includegraphics[width=1.0\textwidth]{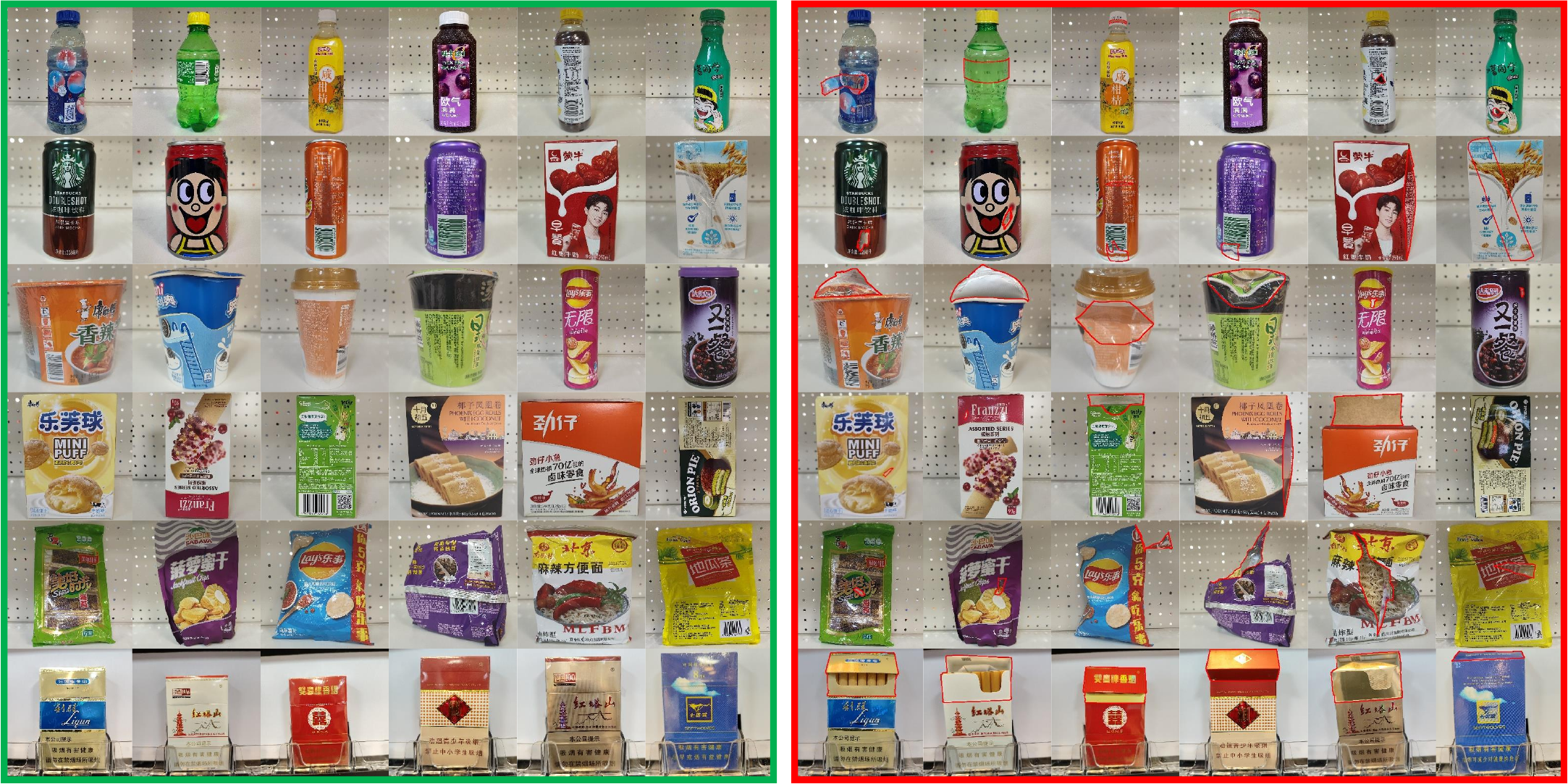}
	\caption{Example images for six categories of the GoodsAD dataset. The first to sixth rows show bottled drink, canned drink, bottled food, boxed food, packaged food and boxed cigarette, respectively. Each category contains multiple items. The green pane on the left shows anomaly-free images, and the red pane on the right shows corresponding anomalous examples. Anomalous regions are highlighted in red polygons.}
	\label{vis}
\end{figure*}
In actual industrial manufacturing, the vast majority of products are normal samples, while anomalous samples only account for a very small number. Therefore in 2019, P. Bergmann et al. proposed an industrial dataset called MVTec AD using one-class classification setting, which means that only normal samples are used during the training phase. This setting is more in line with industrial scenarios and is called semi-supervised or unsupervised anomaly detection. MVTec AD contains 10 object and 5 texture categories with a total of 5354 images. Each image in the dataset contains only one object, and the camera is perpendicular to the object with the same shooting angle. This dataset has drawn a lot of attention and many methods focus on unsupervised anomaly detection based on this dataset. Since then, some datasets \cite{mishra2021vt,jezek2021deep,zou2022spot} have been proposed, using the same settings. BTAD \cite{mishra2021vt} contains 2830 images with 3 different classes (industrial products), of which 1799 anomaly-free images are for training and the rest for testing. Compared with MVTec AD, the shooting conditions of the images in MPDD \cite{jezek2021deep} are more complex. Under different light intensities and non-homogeneous backgrounds, the image captured by the camera contains multiple objects with different spatial directions, positions and distances. VisA \cite{zou2022spot} is a newly proposed dataset with multiple objects in the image, and the number of images is about twice that of MVTec AD.
\par
However, The current datasets contain at most a dozen classes of objects. There is no goods anomaly detection dataset, which is needed in unmanned supermarkets and commodity production. Different types of datasets are also needed in anomaly detection research to test the universality of current state-of-the-art methods and promote real-world applications.

\section{DATASET DESCRIPTION}
\subsection{Problem Statement and Definition}
In practical applications, commodity anomalies are difficult to define in advance for supervised learning, and it is easy to acquire normal samples but costly and limited to get anomalous sample data. Therefore, GoodsAD adopts the same unsupervised setting as the previous datasets \cite{bergmann2019mvtec,zou2022spot}. The training set contains only images without defects. The test set contains both: images containing various types of defects and defect-free images.\par
VAD consists of two sub-tasks, image-level anomaly detection (classification) and pixel-level anomaly localization (segmentation). The input is an image $ I \in \mathbb R^{H\times W \times 3} $, and the output is an anomaly score $ \eta \in [0,1]$ for anomaly classification or a segmentation mask $ M \in \mathbb R^{H\times W}$ for anomaly segmentation. The value range of each pixel of $ M $ is $ [0,1] $, indicating the degree of anomaly.
\subsection{Dataset Details}
The GoodsAD dataset comprises 6 categories with 3136 images for training and 2988 images for testing. Table \ref{tab:ove} gives an overview for each category. Fig. \ref{vis} shows example images for every category together with example defects. We collected 6 kinds of common commodities in supermarkets, which are \textit{drink\_bottle (d\_b)}, \textit{drink\_can (d\_c)}, \textit{food\_bottle (f\_bt)}, \textit{food\_box (f\_bx)}, \textit{food\_package (f\_p)} and \textit{cigarette\_box (c\_b)}. Each commodity can be used and evaluated individually if necessary. Each category contains multiple goods, and the dataset contains a total of 484 goods. As a result, The appearance of each item varies greatly, such as variations in colour and texture. Each category contains several common defects such as surface damage, deformation and opened. The defects contain both surface texture changes and structure changes. The defects were manually generated to produce realistic anomalies as they would occur in real-world application scenarios.
\par
\begin{table}[t]
\centering
\setlength\tabcolsep{2.6pt}
\caption{Overview of GoodsAD dataset}
\begin{threeparttable}
\begin{tabular}{l|c|c|c:c|c:c|c:c|c|c}
\hline
\multirow{2}{*}{Category} & Train & \multicolumn{7}{c|}{Test}                                       & \multirow{2}{*}{Sum}  & Goods \\ \cline{2-9}
                          & good  & good & \multicolumn{6}{c|}{defective (number + type)}                           &      & types      \\ \hline
drink\_bottle             & 733   & 356  & 273 & s\_d  & 79 & c\_o   & 73 &c\_h\_o & 1514 & 97    \\
drink\_can                & 234   & 147  & 62 &s\_d   & 61 &d & 24 &s\_m  & 528  & 59    \\
food\_bottle              & 1014  & 243  & 156 &s\_d  & 86 &d & 119 &o       & 1618 & 60    \\
food\_box                 & 432   & 146  & 74 &s\_d   & 76 &d & 101 &o       & 829  & 57    \\
food\_package             & 540   & 253  & 144 &s\_a & 86 & o     &    &               & 1023 & 95    \\
cigarette\_box            & 183   & 183  & 246 &o      &    &         &      &                   & 612  & 116   \\ \hline
Sum                       & 3136  & 1328 &                             \multicolumn{6}{c|}{1660}                        & 6124 & 484  \\ \hline
\end{tabular}
\label{tab:ove}
\end{threeparttable}
\begin{tablenotes}
	\footnotesize
	\item s\_d, c\_o, c\_h\_o, s\_a, s\_m, d and o represent surface\_damage, cap\_open, cap\_half\_open, surface\_anomaly, deformation and opened, respectively.
\end{tablenotes}

\end{table}
All images are acquired with 3000 $\times$ 3000 high-resolution. The object locations in the images are not aligned. Most objects are in the center of the images and one image only contains a single object. For each item, we collected multiple images from different angles. For bottled and canned goods, we collected images from different angles around the cylinder. The images were acquired under the illumination conditions of a real supermarket. The appearance of goods may change in texture due to illumination.
The image background is a natural white commodity shelf.
Both image-level and pixel-level annotations are provided.
\par
\begin{figure}[t]
	\centering
	\includegraphics[width=1.0\columnwidth]{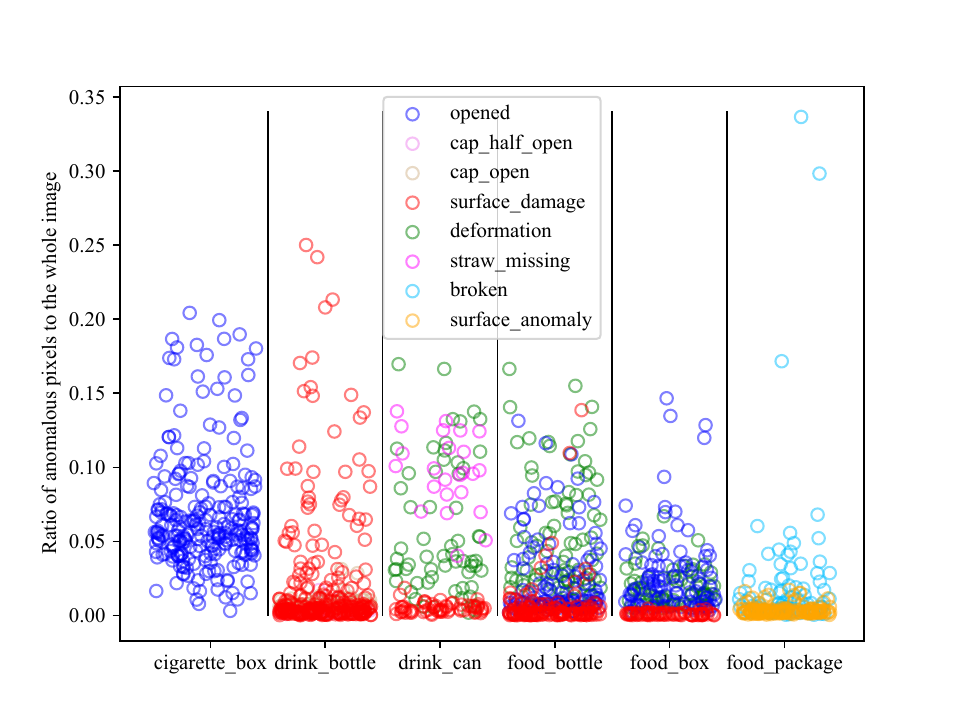}
	\caption{Statistics on the ratio of the number of pixels in the anomalous region to the entire image. Different coloured circles represent different types of anomalous samples.}
	\label{fig:sta}
\end{figure}

Fig. \ref{fig:sta} shows the region size of different anomalies in six categories. Different types of anomalies differ in size, and anomalies of the same type change in size. Most anomalies like \textit{surface damage} and \textit{cap open} occupy only a small fraction (less than 2\%) of image pixels. \textit{opened} and \textit{deformation} are two kinds of anomalies with relatively large proportion.

\section{BENCHMARK}
\subsection{Methods for Visual Anomaly Detection}
Different types of unsupervised SOTA VAD methods are tested on the proposed GoodsAD dataset. We divide current methods into three categories: based on pre-trained models, based on pseudo-anomaly, and based on generative models. Pseudo anomaly-based methods adopt contrastive learning \cite{chen2020simple} paradigms or auto-encoders \cite{kingma2013auto} for image reconstruction. Generative Adversarial Networks (GAN) \cite{goodfellow2014generative}, Normalizing Flow \cite{kingma2018glow} and Diffusion Model \cite{songscore} are the most commonly used generative models, which can be used in VAD. \par
\subsubsection{ Based on pre-trained models} 
This type of approach uses the models pre-trained on ImageNet and does not require a training stage. Because deep learning libraries such as PyTorch provide pre-trained models, it is convenient to use. The basic idea of this type of approach is comparison. We can know whether the test image is anomalous by comparing the test image with the normal training image. Pixel-level comparisons at the image level show that the detection results are too sensitive to pixel values, and there are problems with misalignment of objects. Therefore, Niv Cohen and Yedid Hoshen first proposed the method based on the pre-trained model, SPADE \cite{cohen2020sub}. They used ResNet \cite{he2016deep} to extract the features of the images and compare the feature vectors at the image and patch level. K-Nearest Neighbors (KNN) algorithm is adopted to obtain more robust results. 
\par
PaDiM \cite{defard2021padim} improves SPADE, assuming that the distribution of patches of normal images is subject to multivariate Gaussian distribution, and estimates the mean and variance in the training stage. In the test stage, the Mahalanobis distance between the feature vector of the test image and the distribution is calculated as the anomaly score. PatchCore \cite{roth2022towards} uses greedy coreset subsampling to reduce the memory bank of the normal samples. It uses the second and third level feature maps extracted by Convolutional Neural Network (CNN) such as WideResNet \cite{ZagoruykoK16} and average pooling is adopted on these feature maps to obtain global information. SimpleNet \cite{liu2023simplenet} adopts a simple network architecture and combines the ideas of the pre-trained model and pseudo-anomaly. It improves PatchCore by adding Gaussian noise in feature space and a discriminator. 
\par
The methods based on knowledge distillation \cite{44873} assume that the teacher network and the student network will output different feature maps for anomalous samples in the test stage. MKD \cite{salehi2021multiresolution} uses a smaller student network and multilayer feature synthesis. RD4AD \cite{deng2022anomaly} proposes reverse distillation paradigm and uses the residule block of ResNet to limit the features acquired by the student network.
\begin{table}[t]
\centering
\scriptsize
\caption{Implementation Details of VAD Methods}
\setlength\tabcolsep{0.8pt}
\begin{tabular}{l|c|c|c|c|c}
\hline
    Method    & backbone  & epoch & bs & lr & code \\ \hline
DRAEM \cite{zavrtanik2021draem}    & U-Net &  700    &     8       & 0.0001 & github.com/VitjanZ/DRAEM    \\
RD4AD \cite{deng2022anomaly}    & WideResNet50  &  200    &  16   &  0.005     &  github.com/hq-deng/RD4AD    \\
CutPaste \cite{li2021cutpaste} & ResNet18  & 300    &   32     &  0.03  &    github.com/LilitYolyan/CutPaste  \\
NSA \cite{schluter2022natural}      & ResNet18 &    320  &     64   &  0.001  &   {\tiny github.com/hmsch/natural-synthetic-anomalies}   \\
CFLOW-{\tiny AD \cite{GudovskiyIK22}} & WideResNet50 &  25   &     32    &   0.0002 &   github.com/gudovskiy/cflow-ad   \\ 
SimpleNet \cite{liu2023simplenet} & WideResNet50 & 50& 8 & 0.001 & github.com/DonaldRR/SimpleNet\\
f-AnoGAN \cite{schlegl2019f}    & - &  -   &     32    & 0.0002  &   github.com/A03ki/f-AnoGAN   \\
SPADE \cite{cohen2020sub}    & WideResNet50 &   -   &     -    & -  &   {\tiny github.com/byungjae89/SPADE-pytorch}   \\
PatchCore \cite{roth2022towards}    & ensemble  &  -   &     -    & -  &   {\tiny github.com/amazon-research/patchcore-inspection}   \\
\hline
\end{tabular}
\label{tab:app}
\end{table}
\subsubsection{ Based on pseudo-anomaly}
This type of method simulates natural anomalies to generate some pseudo anomalies in the training phase, so the unsupervised task is transformed into a supervised task. Contrastive learning based methods including CutPaste \cite{li2021cutpaste}, NSA \cite{schluter2022natural} and SPD \cite{zou2022spot} introduce the idea and classical methods of contrastive learning into VAD. The classical contrastive learning method aims to learn the general features of images, while the VAD task needs to detect anomalous areas in the images, so the classical method needs to be modified to adapt to this task. CutPaste cuts an image patch with colour jitter and pastes it at a random location of a large image to generate the anomalous sample. In the training stage, it uses anomaly classification as the proxy task. NSA extracts foreground objects before cutting the image patch and uses Poisson image editing approach to fuse the image patch. 
\par
Image reconstruction based methods such as RIAD \cite{zavrtanik2021reconstruction}, DRAEM \cite{zavrtanik2021draem} and DSR \cite{zavrtanik2022dsr} use the auto-encoders (U-Net \cite{ronneberger2015u} is used for implementation). RIAD introduced image inpainting into image reconstruction to obtain large reconstruction errors of anomalous samples. DRAEM adds a segmentation network after reconstruction network to obtain more accurate results. DSR adopts quantized feature space and moves the anomaly generation process into the feature space. CRDN \cite{10187674} improves DRAEM by cascade network architecture and structural anomaly generation. MemAE \cite{gong2019memorizing} also uses an auto-encoder, but an innovative memory module is adopted to handle the problem of good generalization of the anomalous regions.
\begin{table}[t]
\setlength\tabcolsep{3pt}
\centering
\caption{Performance of image-level anomaly classification}
\begin{threeparttable}
\begin{tabular}{l|l|cccccc|c}
\hline
                      Method &  Metric      & c\_b & d\_b & d\_c & f\_bt & f\_bx & f\_p & avg. \\ \hline \hline
\multirow{2}{*}{CutPaste \cite{li2021cutpaste}}       &    AUROC    &       77.9        &      54.1        &  53.3         &       61.7      &       55.8   &        58.3      &  60.2    \\
                           & AUPR       &       79.1        &      59.2        &        53.4   &    66.4         &        64.9  &         54.0     &   62.8   \\ \hline
\multirow{2}{*}{f-AnoGAN \cite{schlegl2019f}}  &    AUROC    &    86.5           &     54.3         &     51.4      &        67.4     &     59.6     &      57.8        &     62.8 \\
                           &  AUPR      &  87.6             &   55.0           &    52.7       &   75.5          &      70.7    &     57.9         &     66.6 \\ \hline
\multirow{2}{*}{SPADE \cite{cohen2020sub}}     &    AUROC    &        75.6       &     57.3         &    60.7       &       73.3      &     60.2     &        57.4      &     64.1 \\ 
                           &   AUPR     &    79.6           &     61.6         &    64.0       &      79.8       &   71.0       &  56.4 & 68.7 \\
                       \hline
\multirow{2}{*}{DRAEM \cite{zavrtanik2021draem}} & AUROC & 94.5          & 65.7         & 77.4      & 77.7        & 65.0     & 55.6         & 65.9 \\
                       & AUPR  & 95.8          & 67.9         & 75.3      & 83.9        & 75.7     & 53.3         & 71.0 \\ \hline
\multirow{2}{*}{RD4AD \cite{deng2022anomaly}} & AUROC & 65.1          & 67.2         & 63.3      & 76.6        & 63.3     & 63.5         & 66.5 \\
                       & AUPR  & 66.4          & 69.2         & 60.7      & 81.3        & 71.2     & 60.5   & 68.2 \\
                       \hline
\multirow{2}{*}{NSA \cite{schluter2022natural}}       &  AUROC      &    84.8           &      54.7        &      72.3     &      66.6       &   66.1       &        59.1      &    67.3  \\
                           &     AUPR   &      88.9         &       60.3       &      73.0     &  78.5           &    72.9      &       57.0       &   71.8   \\ \hline
\multirow{2}{*}{CFLOW-AD \cite{GudovskiyIK22}}  &   AUROC     &   92.5            &   63.0           &    70.2       &   79.0          &  64.9        &     57.6         &  71.2    \\
                          &   AUPR      &     94.5          & 66.3             &    75.9       &  85.2           & 73.6         &   56.2           &  75.3    \\ \hline
\multirow{2}{*}{SimpleNet \cite{liu2023simplenet}}  &   AUROC     &      97.8         &     70.0        &     70.8      &    78.1         &   72.0      &       63.2       &   75.3  \\
                          &   AUPR      &     98.1          &        73.1      &      75.2     &   86.1          &  77.2        &     62.8        &  78.7   \\ \hline
\multirow{2}{*}{PatchCore{\tiny -1\%} \cite{roth2022towards}} & AUROC & 98.8          & 77.2         & 86.3      & 86.3        & 72.7     & 66.9         & 81.4 \\
                           & AUPR  & 98.1          & 80.0         & 89.9      & 91.0        & 75.3     & 65.4         & 83.3 \\ \hline
\multirow{2}{*}{PatchCore{\tiny -100\%} \cite{roth2022towards}}  & AUROC & 99.2          & 81.4         & 93.9      & 90.0        & 74.3     & 74.2         & 85.5 \\
                           & AUPR  &   98.6            & 82.9             & 95.1          &  93.2           &   75.9       &   70.8           &  86.1    \\ \hline
\end{tabular}
\end{threeparttable}

\begin{tablenotes}
	\footnotesize
	\item c\_b, d\_b, d\_c, f\_bt, f\_bx, f\_p represent class cigarette\_box, drink\_bottle, drink\_can, food\_bottle, food\_box and food\_package, respectively.
\end{tablenotes}

\label{tab:cls}
\end{table}
\begin{figure}[t]
	\centering
	\includegraphics[width=1.0\columnwidth]{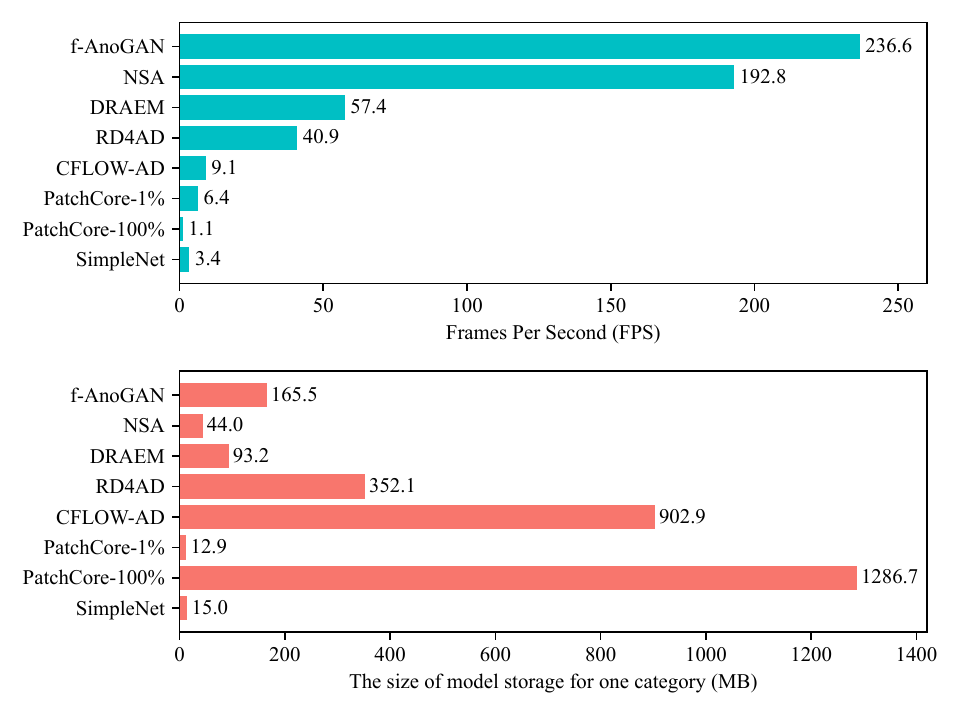}
	\caption{Inference speed and the model parameters stored for each method.}
	\label{fig:fps}
\end{figure}
\subsubsection{ Based on generative models}
The basic idea of this type of method is to use a generative model to fit the distribution of normal samples, and measure the distance between the test sample and the distribution during testing. AnoGAN \cite{schlegl2017unsupervised} introduces GAN into VAD, and the backpropagation algorithm is needed to find the sample closest to the test sample in the distribution. f-AnoGAN \cite{schlegl2019f} solves the problem of slow testing. The method trains a WGAN \cite{arjovsky2017wasserstein} in the first stage, which is the same as AnoGAN, and trains an encoder in the second stage to find the latent encodings of the test sample. CFLOW-AD \cite{GudovskiyIK22} adopts Normalizing Flow to fit the distribution of features extracted by CNN from normal samples. It differs from PaDiM by using a different model to fit the distribution. AnoDDPM \cite{wyatt2022anoddpm} uses the DDPM \cite{ho2020denoising}, the basic idea of which is that anomalous images with added noise can be restored to normal images.\par
Some recent works \cite{xie2023pushing,huang2022registration} focus on a more challenging application scenario: only few-shot (less than 8) normal samples are used in the training stage.
\subsection{Evaluation Metric} 
The standard classification metrics AUROC and AUPR are used for image-level anomaly classification and pixel-level anomaly segmentation. AUPR is more sensitive to the datasets of unbalanced categories. PRO \cite{bergmann2020uninformed} is also adopted to balance anomalous areas of different sizes.
\subsection{Implementation Details}
For each method, we follow one-model-per-category learning paradigm and train one model for each category. It is time-consuming and memory-consuming to train a model for each commodity, although the accuracy is higher in this way.
The images are resized to 224$\times$224 during training and test. All experiments are conducted on NVIDIA GTX 1080Ti GPUs with PyTorch 2.0.
For each method, we adopt the default standard parameters. We set \textit{base\_width} and \textit{base\_channels} in reconstructive and discriminative sub-networks of DRAEM to 64 and 32, respectively. For f-AnoGAN, we train 100000 iterations for WGAN and 50000 iterations for the encoder. More details such as batch size (bs) and learning rate (lr) are listed in Table \ref{tab:app}.

\begin{table}[t]
\setlength\tabcolsep{3pt}
\centering
\caption{Performance of pixel-level anomaly segmentation}
\begin{tabular}{l|l|cccccc|c}
\hline
                      Method &   Metric     & c\_b & d\_b & d\_c & f\_bt & f\_bx & f\_p & avg. \\ \hline \hline
\multirow{3}{*}{SPADE \cite{cohen2020sub}}     &   AUROC     &  40.4             &      59.0        &      66.1     &       72.5      &      59.9    &        72.6      &  61.8    \\ 
                           &  AUPR      &     3.1          &     1.4         &    5.0       &      4.2       &     1.7     & 1.9 &  2.9 \\
                           & PRO & 3.2 & 27.9 & 21.6 & 27.9 & 20.1 & 29.6  & 21.7\\
                       \hline
\multirow{3}{*}{f-AnoGAN \cite{schlegl2019f}}  &   AUROC     &     86.7          &      82.6        &   82.2        &     83.6        &       78.1   &        82.1      &  82.6    \\
 & AUPR & 31.0 & 3.5 & 6.8 & 5.6 & 3.0 & 1.9 & 8.6\\
                           &   PRO     &  69.5             &      56.7        &       53.7    &   55.3          &    48.1      &     42.3         &   54.3   \\ \hline
\multirow{3}{*}{NSA \cite{schluter2022natural}}       &  AUROC      &    72.4           &   84.1           &    81.7       &    88.0         &      80.5    &     84.4         &  81.9    \\
                           &   AUPR     &    24.4           &    9.7          &    14.8       &  29.8           &     9.2     &       7.0       &  15.8    \\ 
                           &     PRO   &         52.1      &       48.3       &     64.8      &   54.0          &    54.4      &       53.5       &   54.5   \\ 
                           \hline
\multirow{3}{*}{DRAEM \cite{zavrtanik2021draem}}  
                           & AUROC & 94.9          & 93.0         & 89.3      & 95.0        & 89.2     & 92.6         & 92.3 \\ 
                           & AUPR  & 58.6          & 22.5         & 23.5      & 38.6        & 16.0     & 14.5         & 23.0 \\ 
                           & PRO & 84.6 & 82.6 & 76.8 & 82.3 &67.8 &67.0 & 72.4 \\
                           \hline
\multirow{3}{*}{RD4AD \cite{deng2022anomaly}}     
                           & AUROC & 91.8          & 95.0         & 93.3      & 96.0        & 90.5     & 94.8         & 93.6 \\
                           & AUPR  & 23.7          & 12.5         & 17.4      & 25.4        & 5.8      & 7.7          & 15.4 \\
                           & PRO & 72.7 & 84.2 & 81.9 & 84.3 & 75.9 & 79.4 & 79.7 \\
                       \hline
\multirow{3}{*}{CFLOW-AD \cite{GudovskiyIK22}}  &  AUROC      &      94.6         &   94.2           &   94.9        &   96.1          &     87.6     &     93.9         &  93.6    \\
                           &     AUPR   &   40.4            & 14.7             &     28.4      &  37.5           &      4.8    &  7.2            &  22.2    \\ 
                            &     PRO   &   81.3            &    74.9          &      72.3     &  80.1           &   62.1       &  70.3            &    73.5  \\ \hline
\multirow{3}{*}{SimpleNet \cite{liu2023simplenet}}  &  AUROC      &  96.0    & 90.9  & 84.0 & 95.0 & 85.5 & 85.6 & 89.5   \\
                           &     AUPR   &    54.1          &       20.8      &       18.7    &    40.5         &      7.1   &       5.6       &  24.4    \\ 
                            &     PRO   &    90.3           &    66.5          &      67.0     &    75.0         &      56.4   &          57.8  &    68.8  \\ \hline
\multirow{3}{*}{PatchCore{\tiny -1\%} \cite{roth2022towards}} & AUROC & 98.7          & 98.8         & 92.2      & 98.2        & 96.2     & 97.1         & 97.0 \\
                           & AUPR  & 71.6          & 65.8         & 54.1      & 63.9        & 19.8     & 34.4         & 51.6 \\
                           & PRO    & 93.4          & 84.2         & 78.0      & 87.3        & 68.2     & 71.9         & 80.5 \\ \hline
\multirow{3}{*}{PatchCore{\tiny -100\%} \cite{roth2022towards}} &   AUROC     & 98.9              &    99.2          &    98.0       &   99.1          &     97.0     &    98.4          &  98.4    \\
                           &  AUPR      &  73.0             &  66.4            &  61.1         &   66.2          & 19.8         &  36.0            &  53.8    \\ 
                           & PRO & 94.0 & 90.6 & 91.2 & 93.5 & 83.1 & 86.7 & 89.9 \\ \hline

\end{tabular}
\label{tab:seg}
\end{table}

\begin{figure*}[t]
	\centering
	\includegraphics[width=1.0\textwidth]{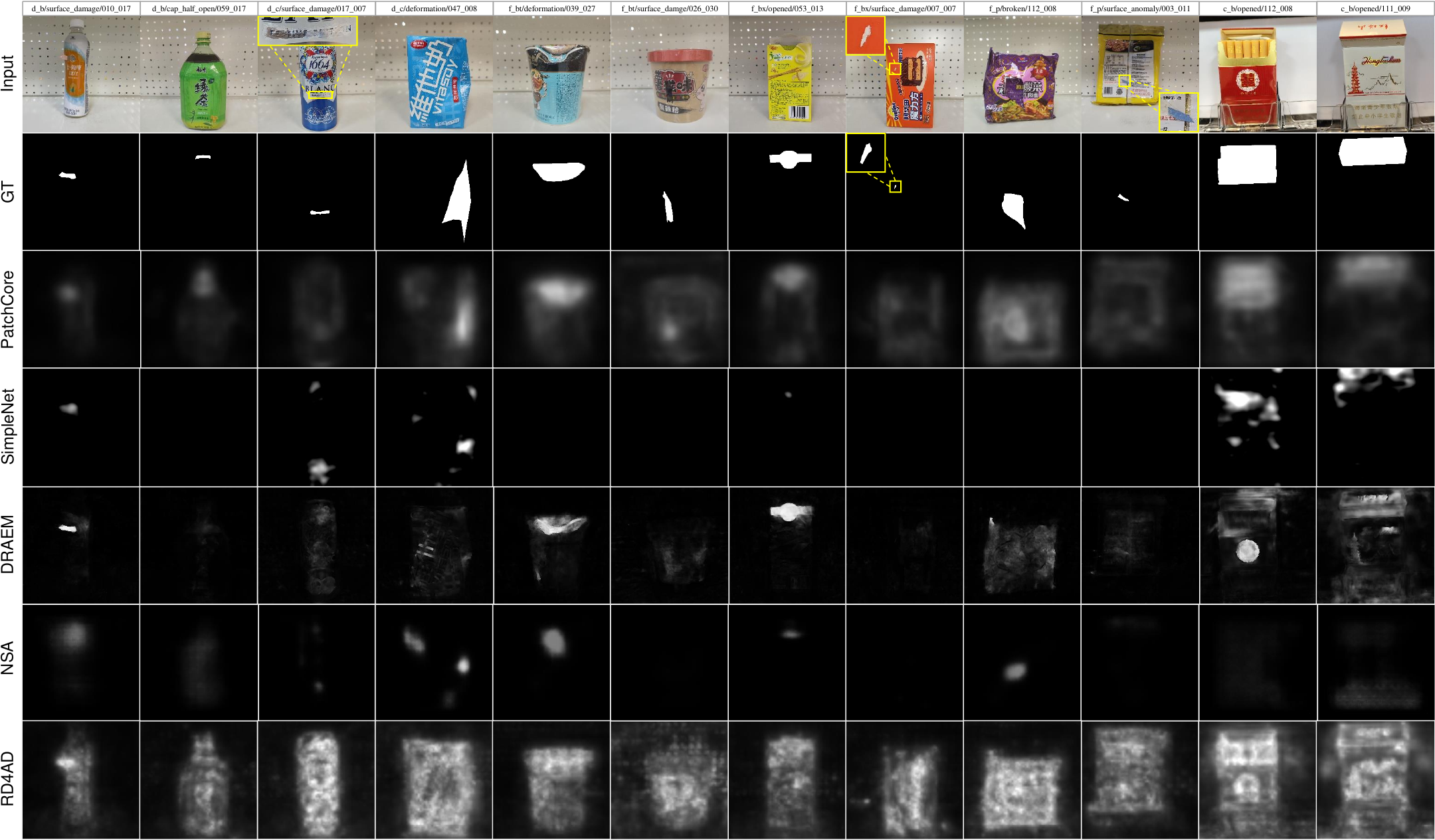}
	\caption{Visual examples of anomaly localization by five different methods on the GoodsAD dataset. GT denotes ground truth. From left to right, each of the two columns is the images of categories drink\_bottle, drink\_can, food\_bottle, food\_box, food\_package and cigarette\_box.}
	\label{fig:vis}
\end{figure*}
\subsection{Experimental Results and Discussion}
We test the performance of different types of methods on the proposed GoodsAD dataset. Table \ref{tab:cls} shows the image-level anomaly classification results and Table \ref{tab:seg} shows the pixel-level anomaly segmentation results. Fig. \ref{fig:vis} shows the qualitative examples of anomaly localization of methods DRAEM, NSA, RD4AD, SimpleNet and PatchCore-100\%.\par
Compared to the previous dataset like MVTec AD, GoodsAD has two different attributes: (1) The object's location in the image is not aligned. (2) The same category contains many items with different appearances. These two characteristics cause the poor performance of current VAD methods. SPADE, RD4AD and CFLOW-AD assume that the location of the object in the image is unchanged, and thus the detection results are not accurate, especially the localization score is low. Because of many goods in one category and appearance change, the student network of RD4AD is challenging to learn the similar representation as the teacher network for normal data samples. Therefore RD4AD incorrectly predicts almost all commodity regions as anomalous, and the samples of anomaly segmentation are shown in the seventh row of Fig. \ref{fig:vis}. RD4AD only achieves 15.4\% AUPR on anomaly localization sub-task.
\par
\begin{figure}[t]
	\centering
	\includegraphics[width=1.0\columnwidth]{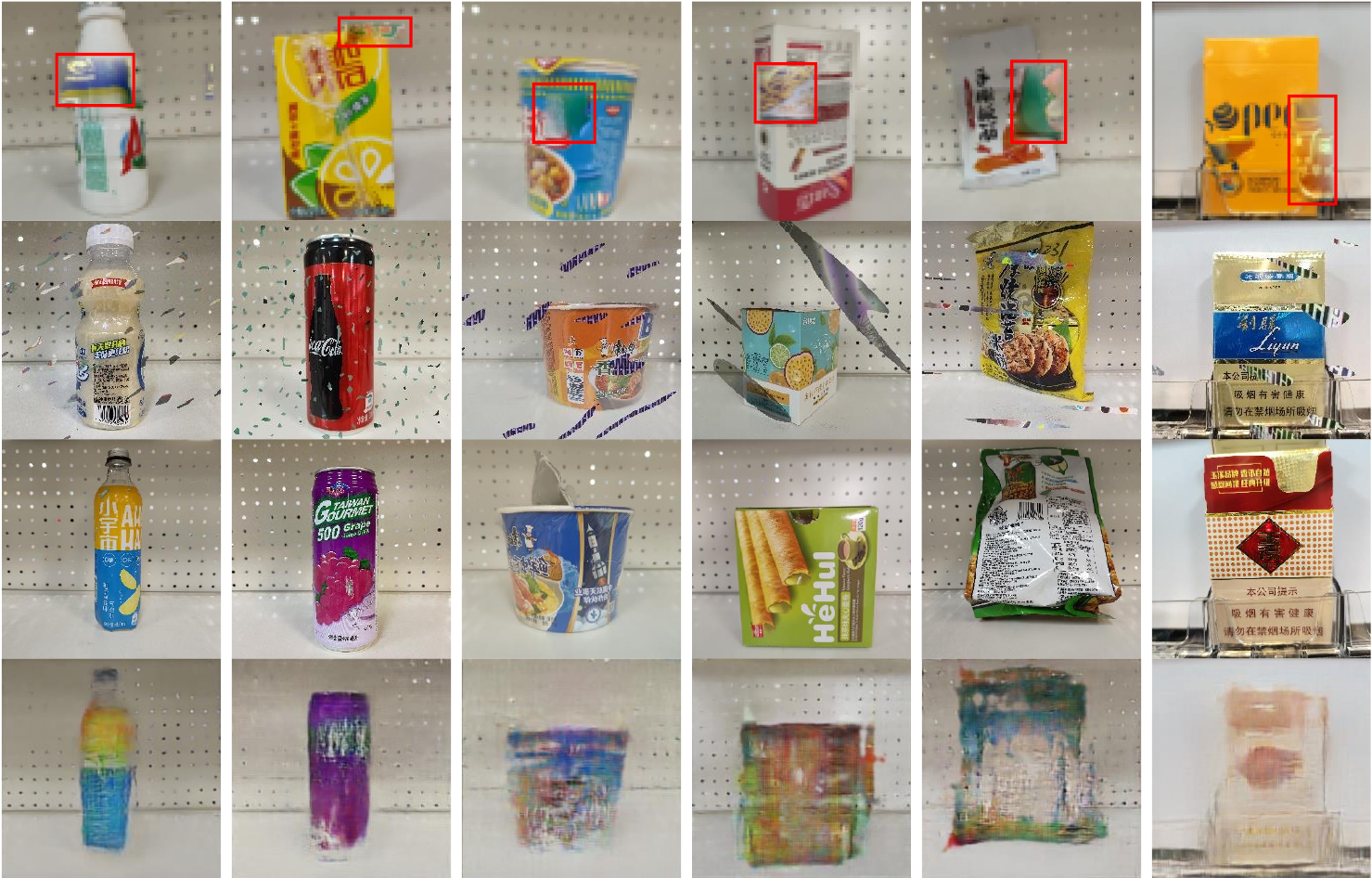}
	\caption{\textit{First row}: Anomaly samples generated by NSA \cite{schluter2022natural}. Anomalous regions are marked with red boxes. \textit{Second row}: Anomaly samples generated by DRAEM \cite{zavrtanik2021draem}. \textit{Third} and \textit{fourth} rows are the anomalous test images from dataset and corresponding normal samples generated by f-AnoGAN \cite{cohen2020sub}.}
	\label{fig:vall}
\end{figure}

The third and fourth rows of Fig. \ref{fig:vall} shows the anomalous test images $x$ and generated normal images $G(E(x))$ by f-AnoGAN \cite{schlegl2019f}. The commodities in the generated images are blurry and the text on the package is not clear. The appearance of the commodity in the generated image mixes up with other commodities (Fig. \ref{fig:vall}, fifth column). Therefore the anomaly segmentation masks obtained by L1 distance $ |x-G(E(x))| $ are not accurate. We think various commodity appearances cause this problem. More training epochs may improve the performance.
\par
As shown in Fig. \ref{fig:vall}, CutPaste and NSA cut a random image patch and blend it into a large image to generate anomalous samples. The generated anomalies are much different from natural anomalies of commodities. Therefore, the detection results of these methods are not accurate, which are shown in the sixth row of Fig. \ref{fig:vis}. NSA only obtains 15.8\% AUPR on the anomaly segmentation task.
\par
Due to the appearance changes and location misalignment of various goods, DRAEM is difficult to learn a proper distance function to recognize the anomaly. The generated samples of pseudo-anomalies in the training stage are shown in Fig. \ref{fig:vall}. DRAEM adopts Perlin noise generator and extra texture images to generate anomalous samples with texture changes. Therefore DRAEM can not recognize anomalies with small texture changes such as bottle cap opening and box deformation (see Fig. \ref{fig:vis}, fifth row). It also fails to detect small anomalies. Apart from PatchCore, DRAEM achieves the second best performance in category cigarette\_box, because the anomalous region \textit{opened} of boxed cigarettes are relatively large and texture changes obviously and the location of boxed cigarettes is relatively aligned.
\par
SimpleNet performs second only to PatchCore on anomaly classification, reaching 75.3\% AUROC. But its anomaly localization score is not high, only getting 24.4\% AUPR. This indicates that SimpleNet can determine whether the image is anomalous but cannot output an accurate anomaly mask. The anomaly masks of boxed cigarettes in Fig. \ref{fig:vis} of SimpleNet are not continuous. It also fails in several samples such as \textit{deformation} of bottled food, \textit{surface\_damage} on boxed and packaged food, and \textit{cap\_half\_open} of bottled drink. We believe that the discriminator of SimpleNet is effective in detecting anomalies but the Gaussian noise is not suitable for commodity anomalies. The accuracy and loss of SimpleNet in the training phase are also unstable. 
\par
From Table \ref{tab:cls} and Table \ref{tab:seg}, PatchCore achieves the best performance among all tested methods. Without subsampling of the memory bank, PatchCore-100\% achieves better performance than PatchCore-1\%. PatchCore-100\% achieves state-of-the-art of 85.5\% AUROC on anomaly classification and 53.8\% AUPR, 89.9\% PRO on anomaly segmentation. PatchCore uses the patch-feature memory bank equally accessible to all patches evaluated at test time, and thus it is less reliant on image alignment. PatchCore adopts KNN algorithm to estimate anomaly scores at test time, and thus it is more robust to the diverse appearance of goods. As shown in sixth row of Fig. \ref{fig:vis}, PatchCore-100\% can predict relatively accurate anomalous regions. Nevertheless, the disadvantage of PatchCore is that the score of the predicted anomalous regions is not high, because image patches sometimes contain both normal and anomalous pixels. PatchCore cannot predict segmentation masks with sharp edges and high confidence like DRAEM (Fig. \ref{fig:vis}, fifth row, first and seventh column). In the ninth column of Fig. \ref{fig:vis}, PatchCore and DRAEM predict normal regions of packaged goods as anomalies due to changes in texture and illumination.
\par
In order to comprehensively evaluate each method, we also test the inference speed and storage space. The results are shown in Fig. \ref{fig:fps}. f-AnoGAN is the fastest method and reaches 236.6 FPS. Although the inference speed of f-AnoGAN and NSA is fast, their performance on the GoodsAD dataset is not high. If the pre-trained model of the PyTorch library is not counted, PatchCore and SimpleNet only need to save the extracted features and the discriminator, respectively. PatchCore-1\% requires less storage space, and its inference speed is 6.4 FPS. PatchCore-100\% requires much space to store the extracted features, but compared to the lightweight PatchCore-1\%, the performance is only slightly improved. CFLOW-AD also occupies much storage space.
\par

From Table \ref{tab:seg}, The scores of the AUROC metric are very high, with most methods exceeding 90\%, but the actual detection results are not accurate. The reason is that the anomalies occupy only a small fraction of image pixels (Fig. \ref{fig:sta}), and the categories of normal and anomalous pixels are extremely unbalanced. The scores of PRO metric are also high. Table \ref{tab:cls} and \ref{tab:seg} show that most methods perform well on category cigarette\_box and the accuracy of food\_box and food\_package is lowest.
\par
In general, current VAD methods do not perform well on the GoodsAD dataset. For real supermarket application scenarios containing a large number of goods, the current methods are not accurate enough for practical application.

\section{CONCLUSION}
In this work, we introduce the GoodsAD dataset, a novel dataset for unsupervised anomaly detection mimicking real-world supermarkets and industrial inspection scenarios. The dataset provides the possibility to evaluate unsupervised anomaly detection methods on a variety of goods with various appearances and different types of anomalies. Pixel-precise ground truth labels are provided to evaluate both image-level classification and pixel-level segmentation. Several current state-of-the-art methods are thoroughly evaluated on this dataset. The best-performing method for all categories is PatchCore. The evaluations show that current methods are not accurate enough for goods anomaly detection and there is still considerable room for improvement. We hope that the proposed dataset will stimulate the development of unmanned supermarkets and smart manufacturing.

\bibliographystyle{IEEEtran} 
\bibliography{IEEEabrv,ref}

\begin{thebibliography}{10}
\providecommand{\url}[1]{#1}
\csname url@rmstyle\endcsname
\providecommand{\newblock}{\relax}
\providecommand{\bibinfo}[2]{#2}
\providecommand\BIBentrySTDinterwordspacing{\spaceskip=0pt\relax}
\providecommand\BIBentryALTinterwordstretchfactor{4}
\providecommand\BIBentryALTinterwordspacing{\spaceskip=\fontdimen2\font plus
\BIBentryALTinterwordstretchfactor\fontdimen3\font minus
  \fontdimen4\font\relax}
\providecommand\BIBforeignlanguage[2]{{%
\expandafter\ifx\csname l@#1\endcsname\relax
\typeout{** WARNING: IEEEtran.bst: No hyphenation pattern has been}%
\typeout{** loaded for the language `#1'. Using the pattern for}%
\typeout{** the default language instead.}%
\else
\language=\csname l@#1\endcsname
\fi
#2}}

\bibitem{bergmann2019mvtec}
P.~Bergmann, M.~Fauser, D.~Sattlegger, and C.~Steger, ``{MVTec AD}--a
  comprehensive real-world dataset for unsupervised anomaly detection,'' in
  \emph{Proc. IEEE Comput. Soc. Conf. Comput. Vision Pattern Recognit.}, 2019,
  pp. 9592--9600.

\bibitem{kermany2018identifying}
D.~S. Kermany, M.~Goldbaum, W.~Cai, C.~C. Valentim, H.~Liang, S.~L. Baxter,
  A.~McKeown, G.~Yang, X.~Wu, F.~Yan, \emph{et~al.}, ``Identifying medical
  diagnoses and treatable diseases by image-based deep learning,'' \emph{cell},
  vol. 172, no.~5, pp. 1122--1131, 2018.

\bibitem{luo2017revisit}
W.~Luo, W.~Liu, and S.~Gao, ``A revisit of sparse coding based anomaly
  detection in stacked rnn framework,'' in \emph{Proc. IEEE Int. Conf. Comput.
  Vision}, 2017, pp. 341--349.

\bibitem{ban2022pdd}
M.~Ban, R.~Ding, J.~Zhang, T.~Guo, and T.~Wang, ``{PDD-Net}: A precise defect
  detection network based on point set representation,'' in \emph{IEEE
  International Conference on Acoustics, Speech and Signal Processing}, 2022,
  pp. 2350--2354.

\bibitem{bozcan2021gridnet}
I.~Bozcan, J.~Le~Fevre, H.~X. Pham, and E.~Kayacan, ``{GridNet}: Image-agnostic
  conditional anomaly detection for indoor surveillance,'' \emph{IEEE Robot.
  Autom.}, vol.~6, no.~2, pp. 1638--1645, 2021.

\bibitem{irvin2019chexpert}
J.~Irvin, P.~Rajpurkar, M.~Ko, Y.~Yu, S.~Ciurea-Ilcus, C.~Chute, H.~Marklund,
  B.~Haghgoo, R.~Ball, K.~Shpanskaya, \emph{et~al.}, ``Chexpert: A large chest
  radiograph dataset with uncertainty labels and expert comparison,'' in
  \emph{Proceedings of the AAAI conference on artificial intelligence},
  vol.~33, no.~01, 2019, pp. 590--597.

\bibitem{lu2013abnormal}
C.~Lu, J.~Shi, and J.~Jia, ``Abnormal event detection at 150 fps in matlab,''
  in \emph{Proc. IEEE Int. Conf. Comput. Vision}, 2013, pp. 2720--2727.

\bibitem{xie2023iad}
G.~Xie, J.~Wang, J.~Liu, J.~Lyu, Y.~Liu, C.~Wang, F.~Zheng, and Y.~Jin,
  ``{IM-IAD}: Industrial image anomaly detection benchmark in manufacturing,''
  \emph{arXiv preprint arXiv:2301.13359}, 2023.

\bibitem{wieler2007weakly}
M.~Wieler and T.~Hahn, ``Weakly supervised learning for industrial optical
  inspection,'' 2007.

\bibitem{song2013noise}
K.~Song and Y.~Yan, ``A noise robust method based on completed local binary
  patterns for hot-rolled steel strip surface defects,'' \emph{Applied Surface
  Science}, vol. 285, pp. 858--864, 2013.

\bibitem{huang2020surface}
Y.~Huang, C.~Qiu, and K.~Yuan, ``Surface defect saliency of magnetic tile,''
  \emph{The Visual Computer}, vol.~36, pp. 85--96, 2020.

\bibitem{zhang2022fdsnet}
J.~Zhang, R.~Ding, M.~Ban, and T.~Guo, ``{FDSNet}: An accurate real-time
  surface defect segmentation network,'' in \emph{IEEE International Conference
  on Acoustics, Speech and Signal Processing}, 2022, pp. 3803--3807.

\bibitem{zou2022spot}
Y.~Zou, J.~Jeong, L.~Pemula, D.~Zhang, and O.~Dabeer, ``Spot-the-difference
  self-supervised pre-training for anomaly detection and segmentation,'' in
  \emph{ECCV}, 2022, pp. 392--408.

\bibitem{jezek2021deep}
S.~Jezek, M.~Jonak, R.~Burget, P.~Dvorak, and M.~Skotak, ``Deep learning-based
  defect detection of metal parts: evaluating current methods in complex
  conditions,'' in \emph{Int. Cong. Ultra Mod.Telecommun. Control Syst.
  Workshops}, 2021, pp. 66--71.

\bibitem{mishra2021vt}
P.~Mishra, R.~Verk, D.~Fornasier, C.~Piciarelli, and G.~L. Foresti, ``{VT-ADL}:
  A vision transformer network for image anomaly detection and localization,''
  in \emph{IEEE Int. Symp. Ind. Electron.}, 2021, pp. 01--06.

\bibitem{tabernik2020segmentation}
D.~Tabernik, S.~{\v{S}}ela, J.~Skvar{\v{c}}, and D.~Sko{\v{c}}aj,
  ``Segmentation-based deep-learning approach for surface-defect detection,''
  \emph{Journal of Intelligent Manufacturing}, vol.~31, no.~3, pp. 759--776,
  2020.

\bibitem{nguyen2022vindr}
H.~Q. Nguyen, K.~Lam, L.~T. Le, H.~H. Pham, D.~Q. Tran, D.~B. Nguyen, D.~D. Le,
  C.~M. Pham, H.~T. Tong, D.~H. Dinh, \emph{et~al.}, ``Vindr-cxr: An open
  dataset of chest x-rays with radiologist’s annotations,'' \emph{Scientific
  Data}, vol.~9, no.~1, p. 429, 2022.

\bibitem{rajpurkar2017mura}
P.~Rajpurkar, J.~Irvin, A.~Bagul, D.~Ding, T.~Duan, H.~Mehta, B.~Yang, K.~Zhu,
  D.~Laird, R.~L. Ball, \emph{et~al.}, ``Mura: Large dataset for abnormality
  detection in musculoskeletal radiographs,'' \emph{arXiv preprint
  arXiv:1712.06957}, 2017.

\bibitem{sultani2018real}
W.~Sultani, C.~Chen, and M.~Shah, ``Real-world anomaly detection in
  surveillance videos,'' in \emph{Proc. IEEE Comput. Soc. Conf. Comput. Vision
  Pattern Recognit.}, 2018, pp. 6479--6488.

\bibitem{chen2020simple}
T.~Chen, S.~Kornblith, M.~Norouzi, and G.~Hinton, ``A simple framework for
  contrastive learning of visual representations,'' in \emph{International
  conference on machine learning}.\hskip 1em plus 0.5em minus 0.4em\relax PMLR,
  2020, pp. 1597--1607.

\bibitem{kingma2013auto}
D.~P. Kingma and M.~Welling, ``Auto-encoding variational bayes,'' in
  \emph{ICLR}, 2014.

\bibitem{goodfellow2014generative}
I.~Goodfellow, J.~Pouget-Abadie, M.~Mirza, B.~Xu, D.~Warde-Farley, S.~Ozair,
  A.~Courville, and Y.~Bengio, ``Generative adversarial nets,'' \emph{Advances
  in Neural Information Processing Systems}, vol.~27, 2014.

\bibitem{kingma2018glow}
D.~P. Kingma and P.~Dhariwal, ``Glow: Generative flow with invertible 1x1
  convolutions,'' \emph{Advances in neural information processing systems},
  vol.~31, 2018.

\bibitem{songscore}
Y.~Song, J.~Sohl-Dickstein, D.~P. Kingma, A.~Kumar, S.~Ermon, and B.~Poole,
  ``Score-based generative modeling through stochastic differential
  equations,'' in \emph{International Conference on Learning Representations}.

\bibitem{cohen2020sub}
N.~Cohen and Y.~Hoshen, ``Sub-image anomaly detection with deep pyramid
  correspondences,'' \emph{CoRR}, vol. abs/2005.02357, 2020.

\bibitem{he2016deep}
K.~He, X.~Zhang, S.~Ren, and J.~Sun, ``Deep residual learning for image
  recognition,'' in \emph{Proc. IEEE Comput. Soc. Conf. Comput. Vision Pattern
  Recognit.}, 2016, pp. 770--778.

\bibitem{defard2021padim}
T.~Defard, A.~Setkov, A.~Loesch, and R.~Audigier, ``{PaDiM}: a patch
  distribution modeling framework for anomaly detection and localization,'' in
  \emph{ICPR}, 2021, pp. 475--489.

\bibitem{roth2022towards}
K.~Roth, L.~Pemula, J.~Zepeda, B.~Sch{\"o}lkopf, T.~Brox, and P.~Gehler,
  ``Towards total recall in industrial anomaly detection,'' in \emph{Proc. IEEE
  Comput. Soc. Conf. Comput. Vision Pattern Recognit.}, 2022, pp.
  14\,318--14\,328.

\bibitem{ZagoruykoK16}
S.~Zagoruyko and N.~Komodakis, ``Wide residual networks,'' in \emph{Proceedings
  of the British Machine Vision Conference}, R.~C. Wilson, E.~R. Hancock, and
  W.~A.~P. Smith, Eds., 2016.

\bibitem{liu2023simplenet}
Z.~Liu, Y.~Zhou, Y.~Xu, and Z.~Wang, ``Simplenet: A simple network for image
  anomaly detection and localization,'' in \emph{Proc. IEEE Comput. Soc. Conf.
  Comput. Vision Pattern Recognit.}, 2023, pp. 20\,402--20\,411.

\bibitem{44873}
G.~Hinton, O.~Vinyals, and J.~Dean, ``Distilling the knowledge in a neural
  network,'' in \emph{NIPSW}, 2015.

\bibitem{salehi2021multiresolution}
M.~Salehi, N.~Sadjadi, S.~Baselizadeh, M.~H. Rohban, and H.~R. Rabiee,
  ``Multiresolution knowledge distillation for anomaly detection,'' in
  \emph{Proc. IEEE Comput. Soc. Conf. Comput. Vision Pattern Recognit.}, 2021,
  pp. 14\,902--14\,912.

\bibitem{deng2022anomaly}
H.~Deng and X.~Li, ``Anomaly detection via reverse distillation from one-class
  embedding,'' in \emph{Proc. IEEE Comput. Soc. Conf. Comput. Vision Pattern
  Recognit.}, 2022, pp. 9737--9746.

\bibitem{zavrtanik2021draem}
V.~Zavrtanik, M.~Kristan, and D.~Skocaj, ``Draem-a discriminatively trained
  reconstruction embedding for surface anomaly detection,'' in \emph{Proc. IEEE
  Int. Conf. Comput. Vision}, 2021, pp. 8330--8339.

\bibitem{li2021cutpaste}
C.-L. Li, K.~Sohn, J.~Yoon, and T.~Pfister, ``Cutpaste: Self-supervised
  learning for anomaly detection and localization,'' in \emph{Proc. IEEE
  Comput. Soc. Conf. Comput. Vision Pattern Recognit.}, 2021, pp. 9664--9674.

\bibitem{schluter2022natural}
H.~M. Schl{\"u}ter, J.~Tan, B.~Hou, and B.~Kainz, ``Natural synthetic anomalies
  for self-supervised anomaly detection and localization,'' in \emph{ECCV},
  2022, pp. 474--489.

\bibitem{GudovskiyIK22}
D.~A. Gudovskiy, S.~Ishizaka, and K.~Kozuka, ``{CFLOW-AD:} real-time
  unsupervised anomaly detection with localization via conditional normalizing
  flows,'' in \emph{Proc. IEEE/CVF Winter Conf. Appl. Comput. Vis.}, 2022, pp.
  1819--1828.

\bibitem{schlegl2019f}
T.~Schlegl, P.~Seeb{\"o}ck, S.~M. Waldstein, G.~Langs, and U.~Schmidt-Erfurth,
  ``f-{AnoGAN}: Fast unsupervised anomaly detection with generative adversarial
  networks,'' \emph{Medical image analysis}, vol.~54, pp. 30--44, 2019.

\bibitem{zavrtanik2021reconstruction}
V.~Zavrtanik, M.~Kristan, and D.~Sko{\v{c}}aj, ``Reconstruction by inpainting
  for visual anomaly detection,'' \emph{Pattern Recogn.}, vol. 112, p. 107706,
  2021.

\bibitem{zavrtanik2022dsr}
V.~Zavrtanik, M.~Kristan, and D.~Skocaj, ``Dsr--a dual subspace re-projection
  network for surface anomaly detection,'' in \emph{European Conference on
  Computer Vision}, 2022, pp. 539--554.

\bibitem{ronneberger2015u}
O.~Ronneberger, P.~Fischer, and T.~Brox, ``{U-Net}: Convolutional networks for
  biomedical image segmentation,'' in \emph{MICCAI}, 2015, pp. 234--241.

\bibitem{10187674}
J.~Zhang, G.~Yang, R.~Ding, and Y.~Li, ``{Cascade RDN}: Towards accurate
  localization in industrial visual anomaly detection with structural anomaly
  generation,'' \emph{IEEE Robot. Autom.}, vol.~8, no.~9, pp. 5560--5567, 2023.

\bibitem{gong2019memorizing}
D.~Gong, L.~Liu, V.~Le, B.~Saha, M.~R. Mansour, S.~Venkatesh, and A.~v.~d.
  Hengel, ``Memorizing normality to detect anomaly: Memory-augmented deep
  autoencoder for unsupervised anomaly detection,'' in \emph{Proc. IEEE Int.
  Conf. Comput. Vision}, 2019, pp. 1705--1714.

\bibitem{schlegl2017unsupervised}
T.~Schlegl, P.~Seeb{\"o}ck, S.~M. Waldstein, U.~Schmidt-Erfurth, and G.~Langs,
  ``Unsupervised anomaly detection with generative adversarial networks to
  guide marker discovery,'' in \emph{Information Processing in Medical
  Imaging}, vol. 10265, 2017, p. 146.

\bibitem{arjovsky2017wasserstein}
M.~Arjovsky, S.~Chintala, and L.~Bottou, ``Wasserstein generative adversarial
  networks,'' in \emph{International conference on machine learning}, 2017, pp.
  214--223.

\bibitem{wyatt2022anoddpm}
J.~Wyatt, A.~Leach, S.~M. Schmon, and C.~G. Willcocks, ``Anoddpm: Anomaly
  detection with denoising diffusion probabilistic models using simplex
  noise,'' in \emph{Proc. IEEE Comput. Soc. Conf. Comput. Vision Pattern
  Recognit. Workshops}, 2022, pp. 650--656.

\bibitem{ho2020denoising}
J.~Ho, A.~Jain, and P.~Abbeel, ``Denoising diffusion probabilistic models,''
  \emph{Advances in Neural Information Processing Systems}, vol.~33, pp.
  6840--6851, 2020.

\bibitem{xie2023pushing}
G.~Xie, J.~Wang, J.~Liu, F.~Zheng, and Y.~Jin, ``Pushing the limits of fewshot
  anomaly detection in industry vision: Graphcore,'' \emph{ICLR}, 2023.

\bibitem{huang2022registration}
C.~Huang, H.~Guan, A.~Jiang, Y.~Zhang, M.~Spratling, and Y.-F. Wang,
  ``Registration based few-shot anomaly detection,'' in \emph{ECCV}, 2022, pp.
  303--319.

\bibitem{bergmann2020uninformed}
P.~Bergmann, M.~Fauser, D.~Sattlegger, and C.~Steger, ``Uninformed students:
  Student-teacher anomaly detection with discriminative latent embeddings,'' in
  \emph{Proc. IEEE Comput. Soc. Conf. Comput. Vision Pattern Recognit.}, 2020,
  pp. 4183--4192.

\end{thebibliography}


\end{document}